\pdfoutput=1%Para compilar en arXiv
\documentclass[lettersize,journal]{IEEEtran}

\usepackage{amsmath,amsfonts}
\usepackage{array}
\usepackage{textcomp}
\usepackage{stfloats}
\usepackage[hyphens]{url}
\usepackage{verbatim}
\usepackage{graphicx}
\usepackage{cite}
\usepackage{subcaption}
\usepackage[english]{babel}
\usepackage[utf8x]{inputenc}
\usepackage[T1]{fontenc}
\usepackage{float}
\usepackage{xurl}
\usepackage{hyperref}
\usepackage{footmisc}
\usepackage{color}
\usepackage[ruled,linesnumbered]{algorithm2e}
\usepackage{tabularx}
\usepackage{xtab,booktabs}
\usepackage{caption}
\usepackage{algpseudocode}
\usepackage{etoolbox}
\usepackage{enumitem}
\usepackage{tabularx}
\usepackage{pifont}
\usepackage{multirow}

\DeclareMathOperator*{\argmax}{\arg\max}
\DeclareMathOperator*{\argmin}{\arg\min}
\DeclareMathOperator*{\argsort}{\arg sort}
\DeclareMathOperator*{\var}{var}
\DeclareMathOperator*{\mean}{mean}
\DeclareMathOperator*{\std}{std}
\DeclareMathOperator*{\selection}{selection}
\DeclareMathOperator*{\update}{update}
\DeclareMathOperator*{\CI}{CI}

\definecolor{grey}{RGB}{167, 160, 159}

\DeclareUnicodeCharacter{2212}{\ensuremath{-}}%Para compilar en arXiv

\makeatletter% Remove right hand margin in algorithm
\patchcmd{\@algocf@start}% <cmd>
  {-1.5em}% <search>
  {0pt}% <replace>
  {}{}% <success><failure>
\makeatother

\newcommand{\cmark}{\ding{51}}
\newcommand{\xmark}{\ding{55}}

\begin{document}

% For algorithms
\RestyleAlgo{ruled}
\SetKwInput{KwInput}{Input}
\SetKwInput{KwOutput}{Output}

% For captions
\captionsetup[figure]{name=Figure,labelsep=period}
\captionsetup[table]{name=Table,labelsep=period}

\title{Speeding-up Evolutionary Algorithms to Solve Black-Box Optimization Problems}

\author{Judith Echevarrieta, Etor Arza and Aritz Pérez
        % <-this % stops a space
\thanks{Judith Echevarrieta (e-mail: jechevarrieta@bcamath.org), Etor Arza (e-mail: earza@bcamath.org), and Aritz Pérez (e-mail: aperez@bcamath.org) are with the Basque Center for Applied Mathematics (BCAM).}% <-this % stops a space
\thanks{Publication date January 10, 2024.}
\thanks{Digital Object Identifier 10.1109/TEVC.2024.3352450}
}

\markboth{P\MakeLowercase{reprint submitted to} IEEE Transactions on Evolutionary Computation\\
A\MakeLowercase{ccepted article} (A\MakeLowercase{uthor's version})~~~~ DOI: 10.1109/TEVC.2024.3352450}%
{Shell \MakeLowercase{\textit{et al.}}: A Sample Article Using IEEEtran.cls for IEEE Journals}


\maketitle

\begin{abstract}
Population-based evolutionary algorithms are often considered when approaching computationally expensive black-box optimization problems. They employ a selection mechanism to choose the best solutions from a given population after comparing their objective values, which are then used to generate the next population. This iterative process explores the solution space efficiently, leading to improved solutions over time. However, one of the challenges of these algorithms is that they require a large number of evaluations to provide a quality solution, which might be computationally expensive when the evaluation cost is high. In some cases, it is possible to replace the original objective function with a less accurate approximation of lower cost. This introduces a trade-off between the evaluation cost and its accuracy.

In this paper, we propose a technique capable of choosing an appropriate approximate function cost during the execution of the optimization algorithm. The proposal finds the minimum evaluation cost at which the solutions are still properly ranked, and consequently, more evaluations can be computed in the same amount of time with minimal accuracy loss. An experimental section on four very different problems reveals that the proposed approach can reach the same objective value in less than half of the time in certain cases.

\end{abstract}


\section{Introduction}
Optimization algorithms try to find a solution to a problem that maximizes an objective function. They can be classified as \textit{exact} or \textit{heuristic} depending on whether the solution provided by them is guaranteed to be the global optimum or not \cite{noauthor_optimization_2013}. The growth of computational complexity in solving certain optimization problems has emphasized the use of heuristic algorithms \cite{rardin_experimental_2001}, among which the population-based evolutionary algorithms stand out \cite{noauthor_configurable_2014}. These algorithms explore the solution space iteratively, leading to the improvement of the best-found solution over time. Given an initial set of solutions, they choose the most promising ones among them (based on their objective values) to generate the next set of candidate solutions. Consequently, they can find a good solution in a reasonable amount of time, although its optimality is not guaranteed. However, a common challenge of these heuristics is that they require a large number of objective function evaluations to provide a near-optimal solution \cite{chugh_survey_2019-1,tenne_computational_2010}. This is because increasing the number of iterations allows more solutions to be compared, leading to a richer exploration of the solution space and achieving a higher solution quality.

The number of evaluations required to solve an optimization problem is an important issue to consider, and even more so when the computational cost of the explicitly unknown objective function is high. One of the real-world application domains that reflects this situation is engineering design. For example, Koch et al. \cite{koch_statistical_1999} work in a context of a wing configuration design of high-speed civil transport which takes approximately 5 minutes to evaluate a single configuration. Further, Liu et al. \cite{liu_multi-fidelity_2016} state that the models used in photonics and microelectromechanical system require well over 1 hour of simulation time per design \cite{koziel_solving_2014}. Finally, Gu et al. \cite{gu_comparison_2001} affirm that a vehicle crashworthiness analysis takes on average 98 hours for a single evaluation. These problems are part of the so-called \textit{computationally expensive black-box problems}~\cite{shan_survey_2010}.

Relating all the aforementioned, heuristic algorithms present certain drawbacks when used to solve computationally expensive black-box problems. These algorithms are able to provide a quality solution as long as we are willing to assume a high computational cost. One way to solve the problem lies in controlling the cost of the objective function. Reducing the cost of evaluations has the potential to yield higher-quality solutions. By decreasing the evaluation cost, more evaluations can be performed within a given runtime, allowing for a more thorough exploration of the solution space. However, it is crucial to ensure that the cost reduction does not compromise the objective function accuracy, otherwise it might prove difficult to identify the solutions with a good real objective value.

\subsection{Related work: early stopping and function approximation}

\begin{figure}[H]
\centering
\includegraphics[width=\columnwidth]{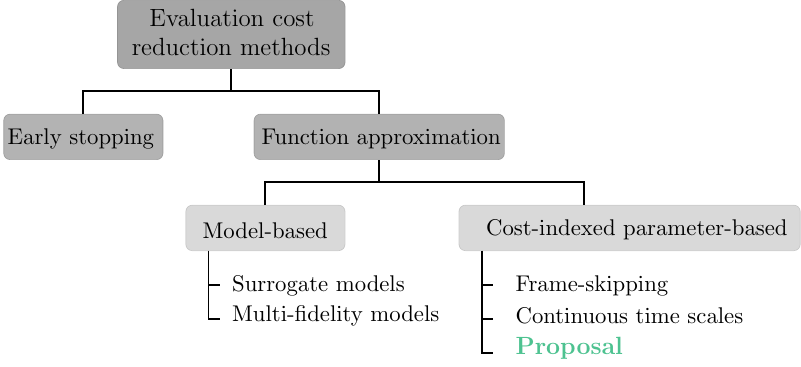} 
\caption{Hierarchical diagram associated with related work.}\label{FIGURE_RelatedWork}
\end{figure}

The search for the balance between cost and accuracy is a problem that has already been approached in the literature by different techniques that reduce the evaluation cost. Some techniques directly limit an evaluation runtime, others instead approximate the objective function to reduce its cost (see Figure \ref{FIGURE_RelatedWork}). 

Techniques of the first type, known as \textit{early stopping}, keep the definition of the objective function intact, but stop the evaluation of a candidate solution when the probability of it performing better than the best known is estimated to be low. These methods have proven successful in hyperparameter optimization \cite{de_souza_capping_2022,hutter_automated_2019,JMLR:v18:16-558}. Moreover, Arza et al. \cite{arza_generalized_nodate-1} recently proposed an adaptation of these methods to work on policy learning tasks, contributing to their generalization. Unfortunately, early stopping methods have a few limitations. Firstly, they are inherently unsuitable for problems with non-monotonous objective functions, such as the target locomotion of an agent in a maze~\cite{arza_generalized_nodate-1,lehman_abandoning_2011}. Additionally, they assume that the objective function can be terminated early obtaining a partial evaluation, which does not apply to some problems such as the Wells turbines design optimization\footnote{The Wells turbines design optimization problem is considered in this work. The objective is to maximize a function that measures the performance of a turbine design and its available functionality is limited to providing the corresponding outputs to given inputs, which makes it impossible to prematurely stop an evaluation (see Section \ref{SECTION_PreliminaryExperiments} for further details).}\cite{zarketa-astigarraga_computationally_2023}.
 
In contrast, the techniques of the second type reduce the cost of an evaluation by modifying the objective function but respecting the evaluation process of a solution without interrupting it. Existing works replace the original objective function with an approximation that consumes less computational time. These techniques are useful in simulation-based optimization problems, commonly found in industry, robotics and engineering. Due to increasing computing capabilities, scenarios in which multiple simulators are available for a given problem are becoming more common, where each simulator has a different fidelity and is associated with an approximate objective function of a certain accuracy~\cite{robinson_multifidelity_2006-1}. However, selecting the level of fidelity of a simulator is not a simple task. For instance, transferring to a robot the policies learned in an imperfect simulation (too low fidelity) may result in poor performance~\cite{cutler_real-world_2015}. Therefore, although lower fidelity simulators are less expensive and allow us to perform more evaluations in a predefined runtime, they offer less accurate results, which may cause a solution quality reduction. 

In the literature about function approximation, the balance to the trade-off between the number of evaluations and the solution quality is tried to find using multi-fidelity simulators. The works can be divided into two groups (see Figure \ref{FIGURE_RelatedWork}): some model the original objective function to obtain approximate functions, while others only modify the value of a strategic parameter to obtain them. Examples of each subgroup are described below.

\subsubsection{Model-based function approximation} The concept of \textit{muti-fidelity} is related to surrogated models  \cite{liu_multi-fidelity_2016,robinson_multifidelity_2006-1}. Optimization methods based on surrogated models minimize costly evaluations of the original high-fidelity objective function by using it in combination with a less expensive but less accurate low-fidelity one. These methods present certain challenges, such as the choice of the supervised model that will replace the computationally costly objective function. The function is approximated by fitting a supervised model from the outputs provided by the original black-box function after evaluating a set of candidate solutions. Normally the type of model (such as polynomial regression (PR), radial basis function (RBF), artificial neural network, kriging or support vector regression) is chosen randomly or because of its popularity in the area with which the problem is associated \cite{chugh_survey_2019-1,diaz-manriquez_selection_2011}. However, although the low-fidelity function provided by the selected surrogate model will be less expensive than the original one, its accuracy may vary depending on the set of solutions used for its adjustment~\cite{lim_generalizing_2010,alizadeh_managing_2020}. Marjavaara et al.~\cite{daniel_marjavaara_hydraulic_2007} use PR and RBF surrogate models in the context of multi-objective optimization and show that each of the models performs better at a different region of the Pareto front. 

The limitation of variable evaluation accuracy of a surrogate model can be avoided with the use of multi-fidelity models. With this method, several low-fidelity models of different complexity and accuracy are available, making it possible to select the most suitable one at each moment of the optimization process. Each of the models has design variables that are defined over different spaces. This means that models of different fidelity are associated with different fixed modelling of the original objective function. While in multi-fidelity models each low-fidelity model does not change during the optimization process, surrogate models are continuously updated during the optimization process. Cutler et al. \cite{cutler_real-world_2015} present a framework for solving policy learning tasks using Reinforcement Learning (RL) algorithms with multi-fidelity simulators that have different state spaces. The framework is designed for an agent to learn in the lowest-level simulator that still provides useful information. It controls the level of fidelity during the policy learning process of a remote control car and specifies the rules for when the agent should switch to a higher or lower fidelity simulator. Robinson et al. \cite{robinson_multifidelity_2006-1} present a methodology that combines two mapping methods defined between different design variable spaces associated with different simulators, allowing the transition between multi-fidelity models and testing its effectiveness on an aircraft wing design problem. 
\subsubsection{Cost-indexed parameter-based function approximation} Remodelling the optimization problem is not the only way to get approximate objective functions of different costs. Some works only modify the value of a strategic parameter to define variable cost simulators. In most of the simulated problems associated with policy learning tasks, agents learn using sequential Markov Decision Processes (MDP) \cite{wiering_reinforcement_2012}. Iteratively, the agent must select an action to issue to the simulator, and the simulator will repeat the action in a given physical time (\textit{time-step}) over several consecutive frames (\textit{frame-skip}). Braylan et al.~\cite{braylan_frame_2015} and Kalyanakrishnan et al.~\cite{kalyanakrishnan_analysis_2021} show that a better performance can be achieved by using a constant frame-skip value well above the original setting in some Atari console games and Classic Control problems such as Acrobot. Furthermore, Sharma et al. \cite{sharma_learning_2017} adapt the value of this parameter online during the lifetime of the agent. Finally, Ni et al. \cite{ni_continuous_nodate} reduce the cost of learning the optimal policy using RL algorithms by keeping the frame-skip at its original value and automatically modifying the time-step parameter during the training process. For this purpose, the dimension of the action space defined in the MDP is increased by one dimension by adding the time-step as an extra continuous action. In this way, when the agent has to make a decision, it will select the action together with the time-step value. 

\subsection{Proposal and contributions}\label{SECTION_ProposalContributions}
This paper presents a novel technique to reduce the cost of an evaluation using approximate objective functions of different costs. Therefore, it is a function approximation method, although in contrast to model-based approximations (such as surrogate models), it is classified in the cost-indexed parameter-based group (see Figure~\ref{FIGURE_RelatedWork}). In contrast to existing works in the literature, we assess the degree of reliability of an approximate function by measuring how well it can rank a set of solutions. In this sense, we define the accuracy of an approximation by ranking the evaluated solutions and comparing them to the ranking of the original function. During the optimization process, the proposed procedure adjusts the optimal evaluation cost of the approximation, which refers to the minimum cost necessary to obtain the desired solution ranking. Three requirements must be fulfilled for the method to be applicable. Firstly, the algorithm used to solve the optimization problem must work with a set of solutions that will be updated iteratively considering their ranking. Secondly, the high evaluation cost must be a bottleneck. Thirdly, the cost of the approximate functions must be controlled and modified by a parameter that is part of the definition of the objective function.

Our proposal overcomes the following three main limitations present in the literature (see Table \ref{TABLE_Contribution}):
\begin{itemize}[wide]
\item[$(i)$]\textit{Model selection}. Surrogate models require a selection of the supervised model type that approximates the original objective function. This selection can be done randomly \cite{chugh_survey_2019-1} or after making an analysis that estimates and compares the overall performance of each type in the problem to be solved \cite{diaz-manriquez_selection_2011}. In any case, by selecting a single model its accuracy is not guaranteed to be the same throughout the entire solution process \cite{alizadeh_managing_2020,lim_generalizing_2010,daniel_marjavaara_hydraulic_2007}, which may lead to a loss in the solution quality. In contrast, our proposal uses a set of approximate functions of different accuracies obtained by modifying a cost-indexed parameter of the original objective function. Therefore, we directly decrease the original objective function complexity, avoiding the selection and training  of a single supervised model and its drawbacks.

\item[$(ii)$] \textit{Adjustment of the approximation cost in real-time.} The proposed technique decides when to adjust the approximation cost during the optimization process, avoiding time and quality loss. Moreover, the cost of readjustment is controlled and its assessment requires no additional evaluations since we recycle previously performed calculations to detect the necessity to update the approximation. In contrast, in surrogate models, for certain instances of the problem it might be possible to find a simpler model that can achieve equally good results as the chosen one \cite{lim_generalizing_2010}. In this situation, the selected model provides an approximation that is still more expensive than necessary, which leads to unnecessary expenditure of time.

\item[$(iii)$] \textit{Problem dependency.} Our technique is not limited to optimization problems associated with policy learning tasks with the presence of parameters of a specific nature (such as frame-skip \cite{braylan_frame_2015,kalyanakrishnan_analysis_2021,sharma_learning_2017} or time-step \cite{ni_continuous_nodate}). The proposal is also tested on other problems of completely different characteristics, some with obvious real industrial applications (the search for the optimal design of a Wells turbine or wind farm layout optimization) and other associated with much more theoretical machine learning concepts (the fitting of a symbolic regression). Therefore, our proposal is more generally applicable, as we show that it works in different problems with very unalike characteristics.
\end{itemize}

\begin{table}[H]
\centering
\begin{tabular}[H]{|l|c|c|c|}
\cline{2-4}
\multicolumn{1}{c|}{}& $(i)$ & $(ii)$  & $(iii)$ \\ 
\hline 
Surrogate models \cite{chugh_survey_2019-1,diaz-manriquez_selection_2011,alizadeh_managing_2020,lim_generalizing_2010,daniel_marjavaara_hydraulic_2007}& \xmark & \xmark  & \cmark \\ 
\hline 
Multi-fidelity models \cite{liu_multi-fidelity_2016,cutler_real-world_2015,robinson_multifidelity_2006-1} & \cmark & \cmark  & \xmark  \\ 
\hline 
Frame-skipping \cite{sharma_learning_2017}& \cmark & \cmark & \xmark  \\ 
\hline 
Continuous time scales \cite{ni_continuous_nodate} & \cmark & \cmark & \xmark  \\ 
\hline 
Proposal & \cmark & \cmark & \cmark   \\ 
\hline 
\end{tabular}
\caption{Summary of the limitations that are overcome by each method described in the literature and by our proposal.}\label{TABLE_Contribution}
\end{table}

In conclusion, with the method proposed in this work, we do not intend to improve the performance of approximate function-based cost reduction methods in the literature, which depends on problem specific properties. Our goal is to define a more general method capable of overcoming all limitations mentioned above. Therefore, to assess the effectiveness of our proposal objectively and generally, in the experimentation we measure the improvement in performance with and without our method.

\subsection{Organization}
The rest of the paper is organized as follows. Section \ref{SECTION_FormalDefinition} introduces the notation, defines the problem, and makes a first formal presentation of the proposed method together with the requirements for its application. Section \ref{SECTION_Method} defines in detail the procedure designed to automatically set the optimal evaluation cost during the execution of the optimization algorithm. Section \ref{SECTION_PreliminaryExperiments} introduces the experimental framework, defining the necessary tools to apply the proposed procedure and testing its possible effectiveness with some initial experiments on the selected problems. Section \ref{SECTION_HeuristicApplication} is the main of the experimentation, where the proposed procedure is applied on the selected problems. Section \ref{SECTION_Limitations} summarizes the limitations of our proposal and Section \ref{SECTION_DisscussionLimitationsFuturework} discusses them indicating some future research directions. Finally, Section~\ref{SECTION_Conclusions} concludes the work.
\section{Problem Definition and Resolution Procedure}\label{SECTION_FormalDefinition}

\begin{figure*}[!t]
\centering
\includegraphics[width=\textwidth]{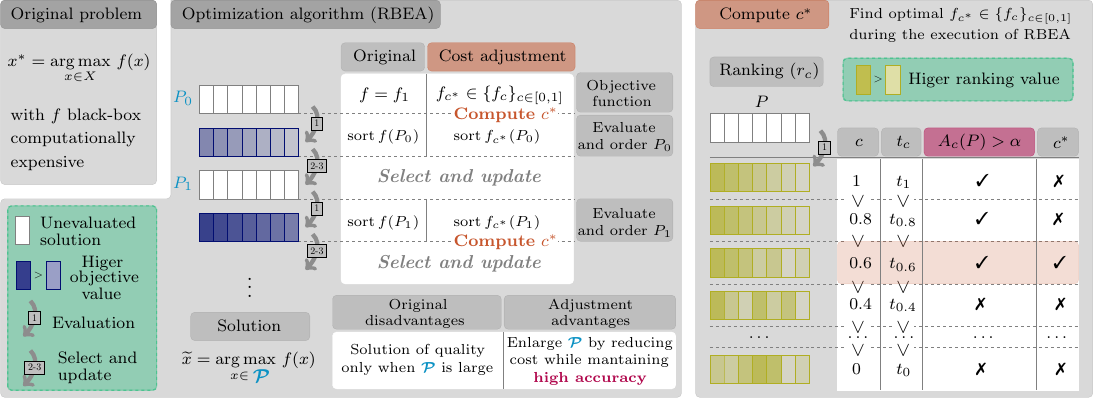} 
\caption{The box on the left (named \textit{Original problem}) defines the problem. The central box (named \textit{Optimization algorithm}) illustrates the original functioning of the RBEA (column \textit{Original}) with its limitations (cell \textit{Original disadvantages}) and compares it with the proposed procedure (column \textit{Cost adjustment}) indicating its advantages (cell \textit{Adjustment advantages}). The box on the right (named \textit{Compute $c^*$}) illustrates the definition of the optimal evaluation cost (defined in Equation~\eqref{EQUATION_OptimalEvaluationCost}).}\label{FIGURE_ResolutionApproach}
\end{figure*}

Solving a \textit{black-box optimization problem} consists of finding the best solution among the feasible ones, which minimizes the cost of a process or maximizes the effectiveness of a system \cite{arora_optimization_2015}. Formally it could be defined as follows\footnote{Without loss of generality, we assume that the optimization problem is a maximization problem. In the case of minimization, it would be enough to replace in Equation \eqref{EQUATION_OptimizationProblem} expression $f(x)$ with $-f(x)$.}
\begin{equation}\label{EQUATION_OptimizationProblem}
x^*=\underset{x \in X}\argmax~ f(x),
\end{equation}
where $X$ is the \textit{solution space} and $f$ is the \textit{computationally expensive objective function} that measures the performance of a solution. Heuristic algorithms provide an approximate solution to an optimization problem. However, when evaluation cost is a bottleneck, the high cost of a single evaluation in certain problems makes even the search for a suboptimal solution very expensive. This section formally presents the fundamentals of the novel method proposed to deal with such a situation, which are schematically illustrated in Figure \ref{FIGURE_ResolutionApproach}.
\subsection{Rank-Based Evolutionary Algorithms}\label{SUBSECTION_OptimizationAlgorithm}
Within the family of heuristic algorithms, in this paper, we focus on population-based evolutionary algorithms, which are characterized by iteratively combining three main procedures: evaluation, selection and updating. We specifically work with those that consider the rankings of the populations, rather than dealing directly with the objective function scores. In these algorithms, the objective function is not directly involved in generating new solutions, its sole purpose is to identify solutions that are more likely to guide the selection process to improved outcomes. Hereafter we refer to this family as \textit{Rank-Based Evolutionary Algorithms} (RBEA), which includes algorithms such as Genetic Algorithms with truncation selection or Estimation of Distribution Algorithms~\cite{simon_evolutionary_2013}. 

The middle box in Figure~\ref{FIGURE_ResolutionApproach} illustrates how RBEAs work originally and their limitations. They iteratively improve a set of solutions by selecting the best-performing subset (using the ranking provided by the objective function) and generating new solutions based on this. They start by generating a random subset of candidate solutions $P_0\subset X$, which represent the solutions of the initial \textit{population}. Then, the \textit{quality} of each solution in the population $P_0$ is evaluated using the objective function $f$ to compare their objective values and make a selection from which the current population will be updated to define a new one $P_1$. This procedure is repeated iteratively until the available computing resources have been exhausted. Given the set of all evaluated solutions $\mathcal{P}$, the solution provided by the RBEA is
\begin{equation}\label{EQUATION_EA_solution}
\widetilde{x}=\underset{x \in \mathcal{P}}{\argmax}~ f(x).
\end{equation}
One of the major challenge of these algorithms is that to ensure that the solution $\widetilde{x}$ of Equation \eqref{EQUATION_EA_solution} is close to the optimal solution $x^*$ of Problem \eqref{EQUATION_OptimizationProblem}, a large number of evaluations have to be executed. Consequently, when the evaluation of a solution is expensive, a high computational cost is required to obtain a quality solution. 

\subsection{Cost-dependent approximations}\label{SUBSECTION_SuitableParameter}
In order to reduce the cost of the optimization algorithm while maintaining the correct ranking of the candidate solutions, we propose to adjust the evaluation cost along the chained evaluation-selection-update process described above (see the columns of adjustments in the middle box of Figure~\ref{FIGURE_ResolutionApproach}). To do this, it is necessary to identify a parameter $\theta$ which takes part in the definition of the objective function $f(x;\theta)$ and whose modification allows us to build approximate functions of different costs and accuracies.

We denote by $c\in[0,1]$ the cost that allows us to index the values of the parameter $\theta$ and abstractly defines the evaluation time of the approximate functions. For each cost $c$ the parameter value $\theta_c$ gives the approximate function $f_c(x):=f(x;\theta_c)$ with an associated evaluation time $t_c$. In this situation, $f_0$ and $f_1$ become the lowest and highest cost approximate function respectively, where $f_1$ is the original objective function (the one defined with the original $\theta_1$ parameter value). We say that the set $\big\lbrace \theta_c\big\rbrace_{c\in[0,1]}$ is a \textit{suitable candidate} for defining approximate functions, if the evaluation times of the associated approximations are linearly related, i.e. if for any $c \in [0,1]$ the evaluation time of $f_c$ is
\begin{equation}\label{EQUATION_Cost}
t_c=t_0+c\cdot(t_1-t_0).
\end{equation}
Note that in this situation, the evaluation time is \textit{monotonically increasing} with the cost, i.e. for any $c,c'\in [0,1]$ such that $c>c'$ it is satisfied that $t_c> t_{c'}$. \footnote{In practice, an approximation of Equation~\eqref{EQUATION_Cost} is considered. However, for the theoretical definition, we use this ideal relation.}

In an iteration of the RBEA, the objective function only affects the selection process through the ranking construction of the current population. Generally, the evaluation of a solution with different approximate functions will be different, i.e. $f_1(x)\neq f_c(x)$ for $c<1$ and $x\in X$. However, this property will not affect the updating process of the RBEA, as long as the selected cost $c\in [0,1]$ provides the same ranking of solutions as the original objective function. Let $r_c$ be the \textit{ranking} induced by the approximate function $f_c$ on the population $P=\lbrace x_1,...,x_{|P|}\rbrace$ formally defined as 
\begin{equation}\label{EQUATION_Ranking}
r_c(P)=\argsort~f_c(P),
\end{equation}
where the operator $\argsort$ builds a tuple with the indexes of the solutions of $P$ after being ordered from highest to lowest according to their corresponding objective values \footnote{In Equation \eqref{EQUATION_Ranking}, we use the notation $f_c(P)$ to refer to the set of the objective values obtained after evaluating the solutions of $P$ with $f_c$, i.e. $f_c(P)~=~\big\lbrace f_c(x)~|~x\in P\big\rbrace$.}. In this situation, we define the \textit{accuracy} of the approximation $f_c$ over the population $P$ as the degree of similarity between the rankings $r_c(P)$ and $r_1(P)$, which we denote by $A_c(P)$. In this work, we propose as accuracy metric Spearman's correlation\footnote{The Spearman's correlation could be replaced by any other metric that assigns the highest accuracy values to the highest cost approximate functions.} defined in $[-1,1]$, where equality $A_c(P)=1$ implies the \textit{ranking preservation}, i.e. $r_c(P)=r_1(P)$. When this equality holds for all populations, departing from the initial population the RBEA obtains the same population sequence as using the original objective function.

The ranking preservation condition is too strict, since no pair of solutions of a population is allowed to be inversely ordered by the approximation. This implies that, in practice, the approximate functions with the maximum accuracy tend to have costs close to one. Consequently, their evaluation times are similar to the time of the original objective function. To ensure time savings in the evaluation process, we propose less restrictive condition setting a threshold to the minimum accuracy of the approximate function. We define the \textit{best approximate function} as the one that finds a balance between cost reduction and solution quality, which is given by the lowest cost approximation that exceeds the accuracy threshold (see the last box in Figure \ref{FIGURE_ResolutionApproach}). Formally, given a population $P$ and a accuracy threshold $\alpha$ the \textit{optimal cost} is defined as follows
\begin{equation}\label{EQUATION_OptimalEvaluationCost}
c^*=\min\big\lbrace c\in[0,1]~|~ A_c(P)>\alpha\big\rbrace.
\end{equation}

In conclusion, the definition of optimal cost makes it easier to select a lower-cost approximation as optimum. Therefore, using $c^*$ to evaluate each population allows saving time while maintaining an accurate enough ranking of solutions such that the RBEA is still able to improve the solution over time.

\section{Efficient Optimal Evaluation Cost Tracking}\label{SECTION_Method}
As mentioned in the previous section, the optimal evaluation cost is the fastest possible approximate function that still properly ranks the solutions. By considering this cost at each iteration of the RBEA, we can reach comparable solution quality as with the original objective function in less time. Furthermore, given a limited computational time, the RBEA with the optimal evaluation cost could discover a solution of higher quality. However, for this to be accomplished, it is necessary to control the computational time involved in computing the optimal cost per population. In this section, we design a procedure to track the optimal evaluation cost efficiently during the execution of the RBEA.

\subsection{Selecting the costs to be evaluated}\label{SUBSECTION_bisection}
The naive approach to select the optimal cost requires computing the accuracy of the approximation $f_c$ over the current population $P$ using all possible cost values $c\in [0,1]$ (see Equation \ref{EQUATION_OptimalEvaluationCost}). However, this is computationally infeasible. By assuming that the accuracy is monotonically increasing with the cost, it is possible to find a cost-effective approximation of the optimal cost using an adaptation of the bisection method~\cite{arora_optimization_2015}.  

Algorithm \ref{ALGORITHM_bisection} shows the considered adapted version of the bisection method, where the cost interval $[0,1]$ that contains the optimal value is progressively narrowed down. In each iteration, the interval is replaced by the half that contains the optimum. We choose the lower or upper half depending on whether the approximation associated with its midpoint is of enough accuracy or not, respectively. As a stopping criterion, we consider reaching an interval length less than 0.1, which is equivalent to performing 4 iterations. In this situation, the bisection allows us to find an approximate cost $\widetilde{c}$ which is within an acceptable range of the optimal cost $c^*$, i.e. \mbox{$|~c^*-\widetilde{c}~|<2^{-4}<0.1$}.

\begin{algorithm}[!h]
\SetAlgoLined
{\small
\caption{Bisection method.}\label{ALGORITHM_bisection}
\KwInput{\\
~~$P\colon$ A set of solutions; \\
~~$\alpha\colon$ Accuracy threshold.}
\vspace{2pt}
\KwOutput{$\widetilde{c}\colon$ Cost-effective approximation of $c^*$.}
$[~c_{low},c_{up}~]\gets [0,1]$\tcp*[f]{Interval containing $c^*$}\;
\For{$c_{up}-c_{low}>0.1$}{
$\widetilde{c}\gets (c_{up}+c_{low})/2$\tcp*[f]{Interval midpoint}\;
	\eIf{$A_{\widetilde{c}}(P)> \alpha$}{
		$[~c_{low},c_{up}~]\gets [~c_{low},\widetilde{c}~]$\tcp*[f]{Lower half}\;}{
		$[~c_{low},c_{up}~]\gets [~\widetilde{c},c_{up}~]$\tcp*[f]{Upper half}\;}
	}
\Return{$\widetilde{c}$}
}
\end{algorithm}

\subsection{Estimating the accuracy of an approximation}\label{SUBSECTION_EstimatingAccuracy}

The described bisection algorithm requires computing the accuracy of four approximations,  which involves evaluating all solutions in a population using different values of cost, including $c=1$. Consequently, computing the approximate optimal cost incurs a higher computational time than evaluating the population with the original function. Therefore, to develop a procedure that speeds up the optimization processes, we propose to estimate the original accuracy value $A_c(P)$ using a random sample of the original population $S \subset P$. In this situation, the time needed to evaluate a population $P$ using the original objective function is
\begin{equation}\label{EQUATION_DefaultPopTime}
t^{original}=|P|\cdot t_1.
\end{equation}
while the maximum time\footnote{The costs $0.5$, $0.75$, $0.875$ and $0.9375$ are those associated with the midpoints of the worst-case bisection when the interval is replaced by its upper half in all iterations. Since the last cost $\widetilde{c}=0.9375$ approximates the real optimum, the time involved in calculating the last accuracy in the bisection is considered as part of the time needed to evaluate the population. Hence, in Equation~\eqref{EQUATION_MaxBisecTime} only the times $t_{0.5}$, $t_{0.75}$ and $t_{0.875}$ participate. } required to compute the approximation of optimal cost is
\begin{equation}\label{EQUATION_MaxBisecTime}
\begin{aligned}
t^{bisec}&=|S|\cdot(t_1+t_{0.5}+t_{0.75}+t_{0.875})\\
&=|S|\cdot (0.875\cdot t_0+3.125\cdot t_1).
\end{aligned}
\end{equation}

We define the sample size $|S|=10$ as the minimum and compensate its value with the frequency of the bisection method application (measured in $t^{period}$ time period), such that the time required for all evaluation cost adjustments constitutes $25\%$ of the available computational time in the worst case. When the ratio between $t^{bisec}$ and $t^{original}$ is above the predefined threshold, the initial sample size is maintained, but the time period with which the method can be applied is recalculated not to exceed the predefined time. Otherwise, the bisection method will be applied with the same time period with which a new population is evaluated using the original objective function, and the sample size can be increased until the allowed time is reached. Specifically, we consider the parameter values shown below.

{\small
\begin{align}
|S|&=
\begin{cases}\label{EQUATION_SampleSize}
 10,  &\text{if}~\dfrac{t^{bisec}}{t^{original}}>0.25\\[2mm]
\bigg\lfloor \dfrac{0.25\cdot t^{original}}{0.875\cdot t_0+3.125\cdot t_1}\bigg\rfloor , &\text{otherwise}
\end{cases}\\[2mm]
t^{period}&=
\begin{cases}\label{EQUATION_TimeFreq}
 4\cdot t^{bisec} , &\text{if}~\dfrac{t^{bisec}}{t^{original}}>0.25\\[5mm]
t^{original}, &\text{otherwise}
\end{cases}
\end{align}}

\subsection{Detecting the need to adjust approximation cost}\label{SUBSECTION_MonitoringVariance}

The naive approach to limit the percentage of the computational budget involved in the optimal cost adjustment procedure to $25\%$ is to apply the bisection method every $t^{period}$ time (defined with Equation~\eqref{EQUATION_TimeFreq}) using the sample size $|S|$ (provided by Equation~\eqref{EQUATION_SampleSize}). However, adjusting the cost at uniform intervals is not necessarily the most efficient approach. This can lead to a loss of time and quality for two reasons. On the one hand, there may be cases for which the period is too small and the optimal cost is recalculated when its value does not vary, leading to unnecessary time expenditure. On the other hand, there may be contrary situations for which the period is too large and the optimum cost is not readjusted when it is needed, so the accuracy of the approximation used might be undesirably low.  To overcome this limitation, we monitor the variance of the population scores during the execution process to decide if we should update the cost, or we should reserve the update for a later time when it is really needed.

As explained in Section \ref{SUBSECTION_SuitableParameter}, we can understand the score provided by an approximate function as a perturbation of the original score (the one obtained with the maximum cost objective function). Intuitively, if the original scores of the solutions forming a population are very different from each other, their perturbed versions can be more easily differentiated and correctly ranked. Otherwise, when the set of original scores is homogeneous, the risk of perturbations affecting the ranks will be greater. Consequently, in the latter case an approximate higher-cost function should be used to evaluate the population and ensure that the selection process of the algorithm does not vary. Therefore, when the variance of the scores of the last evaluated population is significantly lower or higher than the previous ones, this will indicate that the current approximation should be adjusted.

\subsection{Procedure for tracking optimal evaluation cost efficiently}\label{SUBSECTION_OPTECOT}
In the following, we describe in detail the proposed procedure that adjusts the optimal evaluation cost during the optimization process as needed. The pseudocode of the proposed Optimal Evaluation Cost Tracking (\mbox{OPTECOT}) procedure is shown in the Algorithm~\ref{ALGORITHM_heuristic}.

\begin{algorithm}[!h]
\SetAlgoLined
{\small
\caption{Optimal Evaluation Cost Tracking.}\label{ALGORITHM_heuristic}
\KwInput{\\
~$t^{max}$: Maximum time set to execute algorithm; \\
~$t^{period}$: Time period that limits the bisection method cost;\\ 
~$\alpha$: Accuracy threshold;\\
~$\beta$: Number of variances for confidence interval ($\CI$) calculation; \\
~$\kappa$: Number of last optimal costs to assess procedure interruption. }

\vspace{4pt}
$P\gets P_0$\;
	$\widetilde{c}\gets \textup{Algorithm~\ref{ALGORITHM_bisection}}(S\subset P,\alpha)$\;
$B\gets 1$\tcp*[f]{Bisection method applications}\;
$V\gets \emptyset$\tcp*[f]{Set of variances}\;
\While{$t<t^{max}$}{
	
	\If{$|V|>\beta$}{

	$\lbrace v_1,...,v_{\beta},v\rbrace\gets \textup{last }\beta+1\textup{ elements of }V$\; 
	\uIf{$\widetilde{c}=0.9375~\textup{in the last }\kappa~\textup{iterations}$}{
	$\widetilde{c}\gets 1$\;
	}
	\ElseIf{$v\notin\CI\big(\lbrace v_1,...,v_\beta\rbrace\big)~\textup{\textbf{and}}~B<\lfloor t/t^{period}\rfloor$}{
		$\widetilde{c}\gets \textup{Algorithm~\ref{ALGORITHM_bisection}}(S\subset P,\alpha)$\;
		$B\gets B+1$
			}
	}
	$F\gets \lbrace f_{\widetilde{c}}(x)~|~x\in P\rbrace$\tcp*[f]{Evaluate population}\;
		$V\gets V\cup \lbrace \var(F)\rbrace$\tcp*[f]{Add new variance}\;
	$P\gets \update(\selection(P))$\;}
}	
\end{algorithm}

The procedure starts adjusting the optimal evaluation cost, applying the bisection method to the first population (lines 1-2). In the following $\beta+1$ iterations, the populations are evaluated with the optimal evaluation cost approximated in the first iteration. Once this number of iterations has been exceeded, it is possible to start assessing whether the evaluation cost should be readjusted (line 6), using the variances of the sets of scores obtained after evaluating past populations. When the last recorded variance is outside the $95\%$ confidence interval of the previous $\beta$ variances (see Appendix \ref{APPENDIX_CI} for the explicit confidence interval definition), we say that the current optimal cost has become obsolete. In this situation, the evaluation cost will be readjusted to avoid losing quality or time, as long as the time period defined for limiting the evaluation time of applying the bisection method allows it (line 10). The second condition in line 10 ensures that the time spent adjusting the cost does not exceed $25\%$ of the currently elapsed time.

Due to the way RBEAs work, the quality of the solution improves iteratively. As the runtime increases, the populations are formed by solutions of higher quality (all of them closer to the optimum). Therefore, when the algorithm begins to converge, the solutions in the population are very similar to each other and the optimal cost that accurately orders these solutions is very close to the maximum. This implies the execution of the most costly version of the bisection, when it tries to reach the maximum cost by replacing the current interval with its upper half on all iterations, approximating the optimal cost to $0.9375$. Consequently, the application of bisection becomes inefficient: it leads to a loss of quality (use of too low accurate approximation) and time (the same cost is chosen consecutively). To overcome these limitations, we propose to stop adjusting the cost once it has been detected that the last $\kappa$ optimal costs are equal to $0.9375$. Instead, we evaluate each subsequent population with the original objective function (lines 8-9).

\subsection{Guarantee of OPTECOT effectiveness}
We designed \mbox{OPTECOT} such that the optimal cost adjustment will take at most $25\%$ of the available runtime, and by using the optimal cost approximation with high accuracy, time is saved evaluating each solution. Therefore, for the method to be effective, the saved time in the evaluation should compensate for the time expended in bisection and the minimal accuracy losses when using approximate functions. The degree of compensation depends on two factors: the amount of time saved and how much accuracy is lost. These two factors vary depending on the problem, since for the same value of $c$, the ratio $t_1/t_c$ and the value $A_c$ are different. Therefore, the effectiveness of \mbox{OPTECOT} depends on the problem to be solved.

The optimal cost $\widetilde{c}$ defined by \mbox{OPTECOT} during the optimization process is the one that saves evaluation time ($t_{\widetilde{c}}<t_1$) while ensuring a high accuracy ($A_{\widetilde{c}}>\alpha$). Therefore, the effectiveness depends on how large the $t_1/t_{\widetilde{c}}$ ratio is, which represents the number of evaluations that can be performed in the same amount of time needed by the original objective function to evaluate a single solution.  Specifically, the higher this ratio is, the greater the number of solutions that can be evaluated or the iterations of the RBEA that can be executed in the available runtime, which allows a better solution to be found. We further analyze this property in the experimental section (in Section~\ref{SUBSECTION_EffectivenessExperimentation}).

\section{Approximate objective functions for benchmark and real-world problems}\label{SECTION_PreliminaryExperiments}
In this section we present the experimentation. Firstly, the problems selected to apply the \mbox{OPTECOT} procedure are described. Secondly, the approximate objective functions for the procedure application are defined, which are supported by some initial experiments providing intuitions on the use of the approximations in the selected benchmark and real-world problems.

\subsection{Description of the selected optimization problems}
\label{SUBSECTION_EnvironmentDescription}

\begin{figure*}[!t]
\centering
\includegraphics[width=\textwidth]{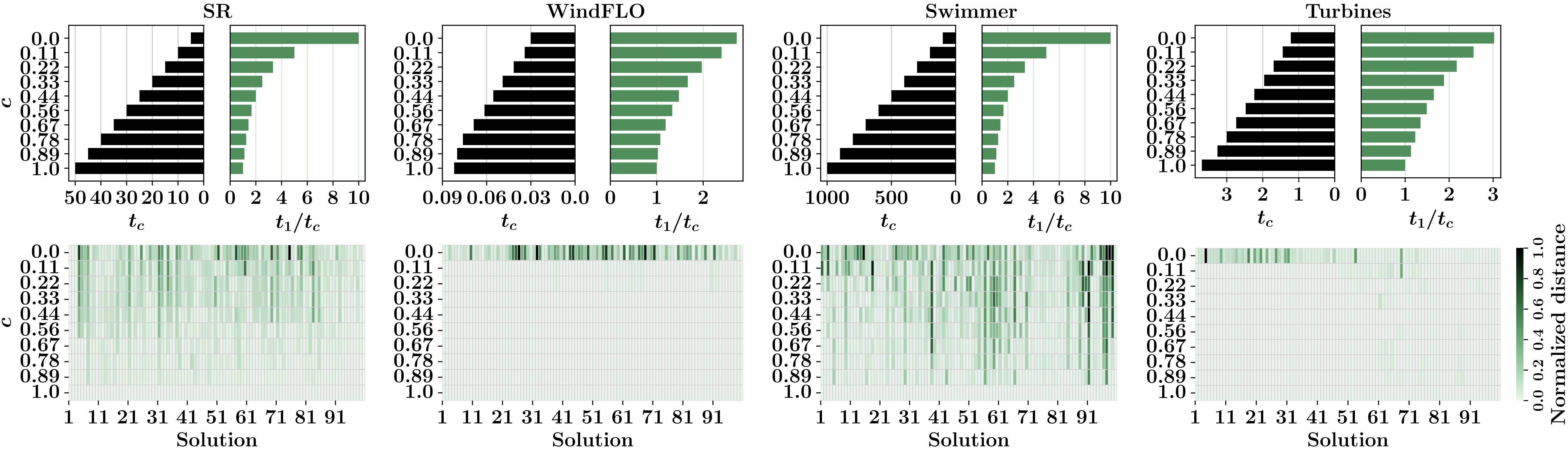} 
\caption{In the first row, the mean evaluation times using different cost values and the proportion of extra solutions that can be evaluated when the cost is decreased. The units of time in each problem are given in Table~\ref{TABLE_TrainTerms}. In the second row, normalized absolute distances between the positions of the approximate rankings and the original.}\label{FIGURE_UnderstandingAccuracyAnalysis}
\end{figure*}

The selected four problems form a heterogeneous group. The first is a regression problem, two are associated with fluid simulations and the last is a policy search problem. They are described below in detail, indicating in each of them which is the optimization algorithms to be used and the cost-indexed parameter (denoted as $\theta$ in Section \ref{SUBSECTION_SuitableParameter}) which allows us to define the approximate functions. 

\textit{Symbolic Regressor (SR).} Symbolic Regression is a machine learning technique that aims to identify an underlying mathematical expression that best describes a relationship of a given set of points. The searched expression is made of a mix of variables, constants and functions, such as addition, subtraction, and multiplication. The optimal mathematical expression is found by a Genetic Programming (GP) algorithm, minimizing the mean absolute error (MEA) \cite{stephens_gplearn_nodate}. This error is calculated by the average distance from the mathematical expression to the set of points. The evaluation cost can be reduced by considering a subset of these points instead of considering all of them.

\textit{WindFLO.} Wind Farm Layout Optimization (WindFLO) \cite{reddy_wind_2020} is a publicly available environment developed to design the layout of wind farms. The optimization problem behind it is finding the optimal layout of a given set of wind turbines on a two-dimensional continuous grid. The optimal solution is the one that maximizes the annual energy production and we find it considering the Covariance Matrix Adaptation Evolution Strategy (CMA-ES) \cite{hansen_cma_nodate} as the optimization algorithm. The energy production is evaluated through the average wake velocity over the rotor-swept area, and is computed with Monte Carlo integration. The number of points that are used to evaluate the integrand is specified by the parameter called \mbox{\textit{monteCarloPoints}}. Reducing the amount of Monte Carlo points reduces the evaluation time but also decreases the accuracy of the evaluation.

\textit{Swimmer.} Swimmer is one of the MuJoCo environments~\cite{farama-foundation_gymnasium_2023}. The swimmers consist of three segments connected by two articulations to form a linear chain. It is suspended in a two-dimensional pool and the goal is to move as fast as possible towards the right by applying torque on the articulations. For each step the swimmer advances to the right, a reward proportional to the distance covered will be assigned, but a cost will also be subtracted if it takes actions that are too large. Finally, an episode is truncated when the maximum length of it is reached. Therefore, in our optimization context, the objective is to find the optimal policy capable of obtaining the maximum possible reward per episode for whatever its initial state is. We use the CMA-ES implementation for MuJoCo environments \cite{sardana_garage_2023} to solve the problem and the parameter time-step to define approximate functions.

\textit{Turbines.} This problem is a real-world problem provided by Mondragon University (MGEP)\cite{zarketa-astigarraga_computationally_2023}. The problem focuses on the design of Wells turbines \cite{wells_fluid_1976}. Each possible design is defined by a total of six variables, such as the number of blades, airfoil distribution and hub ratio. The main objective is to find the turbine design that provides the highest performance in ten different sea states. This optimization problem is solved using the CMA-ES algorithm. The parameter selected to reduce the cost per evaluation is the number of subdivisions on the surface of each blade during simulation, denoted as $N$. Each blade is divided into $N$ small elements and then the force exerted on each of them during one rotation is determined to calculate the total rotor force.

\begin{table}[H]
\centering
\begin{tabular}[H]{|c|c|c|c|c|c|}
\hline 
\textbf{Problem} & $x$ & $f$  & $\theta$ & $\theta_0$ & $\theta_1$ \\ 
\hline 
SR \cite{stephens_gplearn_nodate} & Surface & MAE  &Point set size & 5 & 50 \\ 
\hline 
WindFLO \cite{reddy_windflo_2023} & Layout & Energy & \textit{monteCarloPts} &1&1000 \\ 
\hline 
Swimmer \cite{farama-foundation_gymnasium_2023} & Policy & Reward & Time-step  & 0.1 & 0.01 \\ 
\hline 
Turbines \cite{zarketa-astigarraga_computationally_2023} & Design & Score & $N$ & 10 & 100  \\ 
\hline 
\end{tabular}
\caption{Optimization problems (single solution $x$ and objective function $f$) and cost-indexed parameters (parameter name $\theta$ together with its original $\theta_1$ and lowest-cost $\theta_0$ values).}\label{TABLE_IdentificationTerms}
\end{table}

Table \ref{TABLE_IdentificationTerms} summarizes the characteristics of each problem. It should be noted that the original values of the parameters (denoted by $\theta_1$) are taken from the literature \cite{stephens_gplearn_nodate,reddy_windflo_2023,farama-foundation_gymnasium_2023,zarketa-astigarraga_computationally_2023}.

\subsection{Defining suitable approximate objective function set}\label{SUBSECTION_ResultsSuitability}

To apply the OPTECOT, it is necessary to define a set of
approximate functions whose evaluation times are appropriate
for applying the bisection method. To achieve this goal, we select a set of 10 values for the parameter $\theta$ indexed with equidistant values of cost, i.e. $\lbrace \theta_c\rbrace_{c\in C}$ where $C=\lbrace 0,0.11,...,1\rbrace$, whose associated evaluation times approximately satisfy Equation~\eqref{EQUATION_Cost}~\footnote{The explicit values of the set of parameters and the approximate expression of Equation~\eqref{EQUATION_Cost} that the associated approximate functions fulfill are defined in detail in Appendix~\ref{APPENDIX_RelationCostParamaeter}.}. The approximate functions associated with those parameters are evaluated over a subset of 100 random solutions $\lbrace x_1,x_2,...,x_{100}\rbrace\subset X$, obtaining information about their evaluation times and accuracies.

\subsubsection{Evaluation time of the approximate functions}
The black bar diagrams in the first row of Figure~\ref{FIGURE_UnderstandingAccuracyAnalysis} show the average evaluation times $t_c$ obtained after evaluating the 100 solutions using $f_c$ for all $c\in C$. In addition, the green bar diagrams represent the proportion of extra solutions that can be evaluated in time $t_1$ for each cost. We conclude that evaluation times grow approximately linearly with cost. This allows running more extra evaluations in the same time required by the original objective function. Notice that the shortest evaluation times (shortest black bars) are associated with the largest amount of evaluations (longest green bars). 

\subsubsection{Accuracy of the approximate functions} The second row of Figure~\ref{FIGURE_UnderstandingAccuracyAnalysis} shows the ranking preservation for the different cost values in the four problems. The solutions are sorted according to the original function $f_1$, being $x_1$ and $x_{100}$ the highest and lowest quality solutions, respectively. The color of the square $(i,c)$ represents the normalized difference between the ranking position obtained with $f_1$ and $f_c$ for the solution $x_i$ (a lighter color is associated with a smaller distance), for all $c\in C$ and $i\in\lbrace 1,2,...,100\rbrace$. 

We observe that ranking preservation is fulfilled in all five graphs for the original objective function (the squares of the last rows are of the lightest possible colour). Moreover, as $c$ decreases, the greater the discrepancy between rankings (the squares become darker), which illustrates that the accuracy of the approximate functions decreases with $c$. It follows that it will be difficult to find a cost less than the maximum that rigorously satisfies ranking preservation (except for the graphs associated with WindFLO and Turbines, the second to last rows show squares with a darker colour than the lightest). However, thanks to the accuracy threshold defined in Section \ref{SUBSECTION_SuitableParameter}, we will be able to reduce the evaluation cost as long as the classification in the first positions of the ranking is correct.

\subsubsection{On the potential effectiveness of OPTECOT}\label{SUBSECTION_EffectivenessExperimentation} 
From Figure~\ref{FIGURE_UnderstandingAccuracyAnalysis}, we concluded that for the same cost, the evaluation time saving and the evaluation accuracy vary depending on the problem. However, based on the experiment performed on a sample of $100$ solutions, we can expect that for each selected problem there will be a cost value that saves significant evaluation time and its evaluation accuracy is acceptable.  In Figure~\ref{FIGURE_UnderstandingAccuracyAnalysis} the cost values that could fulfill this statement for Symbolic Regressor, WindFLO, Swimmer and Turbines would be $0.67$, $0.11$, $0.22$, and $0.11$, respectively.

\subsection{Solving the problems with constant evaluation costs}\label{SUBSECTION_ConstantEvaluationCost}
\begin{figure*}[!t]
\centering
\includegraphics[width=\textwidth]{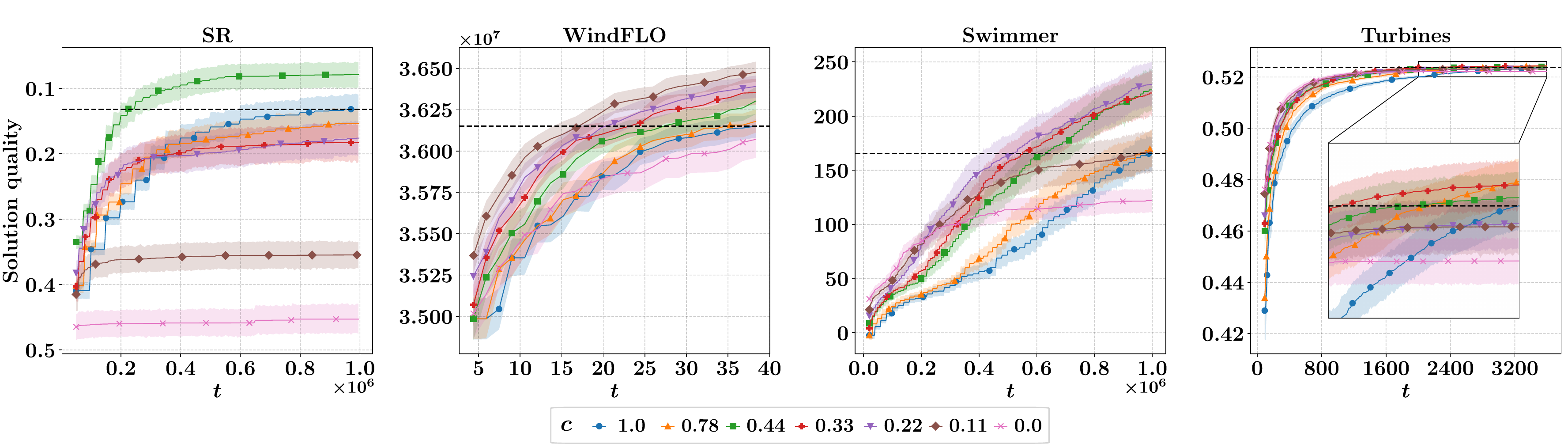} 
\caption{The quality evolution of the best solution found at each runtime for different cost values of the approximate function. The quality is measured in terms of the original objective function and the units of time in each problem are given in Table~\ref{TABLE_TrainTerms}.}\label{FIGURE_ConstantEvaluationCost}
\end{figure*}
In this section we analyze the consequences of using an approximate function during the execution process of the optimization algorithm. We study the behaviour of the solution quality in real-time when using different approximations. This shows the advantages and disadvantages of each approximation, giving us an intuition of the trade-off between cost and quality during the execution. To carry out the experiment, we solve each optimization problem using the $f_c$ approximate function 100 times with different seeds, for all $c\in C$ \footnote{Seven of the 10 costs have been selected in each problem to make the graphical results clearer.}. The runtime limits and population sizes considered in each problem have been taken from the literature\footnote{For Turbines there are no references available on these parameters, hence as a sample size we consider the minimum among the rest of the problems and we define a runtime limit large enough to reach the convergence of the RBEA. } \cite{stephens_gplearn_nodate,reddy_windflo_2023,raffin_ppo_nodate,sardana_garage_2023} and are shown in Table \ref{TABLE_TrainTerms}.
\begin{table}[H]
\centering
{\footnotesize
\begin{tabular}[H]{|c|c|c|c|c|}
\hline 
\textbf{Problem} & $t$ &$t^{max}$ & \textbf{Algorithms} & $|P|$ \\ 
\hline 
SR \cite{stephens_gplearn_nodate} & Points & $10^6$ & GP & 1000\\ 
\hline 
WindFLO \cite{reddy_windflo_2023}& Seconds & 40   & CMA-ES & 50  \\ 
\hline 
Swimmer \cite{raffin_ppo_nodate,sardana_garage_2023} &  Steps & $10^6$ & CMA-ES& 20\\ 
\hline 
Turbines \cite{zarketa-astigarraga_computationally_2023}& Seconds & 3600 & CMA-ES & 20 \\ 
\hline 
\end{tabular} }
\caption{Terms related to train process (measure of runtime $t$, its maximum limit $t^{max}$, optimization algorithm and population size $|P|$).}\label{TABLE_TrainTerms}
\end{table}

Each curve in Figure \ref{FIGURE_ConstantEvaluationCost} associated with a cost represents the average quality of the best solution found by the approximation at each time of the execution process. The quality is measured with the original objective function. Using the notation defined in Section \ref{SECTION_FormalDefinition}, formally the \textit{quality curves} are defined as follows for each cost $c\in[0,1]$
\begin{equation}\label{EQUATION_CurvesConstantAnalysis}
Q_c(t)=f_1\Big(\underset{x\in X_t}\argmax~ f_c(x)\Big), \quad t\in[0,t^{max}]
\end{equation}
where $X_t$ is the set formed by the solutions that define the populations that have been completely evaluated at time $t$. The height of the horizontal dashed line in Figure~\ref{FIGURE_ConstantEvaluationCost} is the \textit{maximum quality} obtained using the original objective function, i.e. $Q_1(t^{max})$. The first coordinate of the points where each curve intersects the line defines the \textit{time to reach the maximum quality} with each approximation denoted as 
\begin{equation}
t_c^{best}=\max\big\lbrace t\in[0,t^{max}]~|~ Q_c(t)\leq Q_1(t^{max})\big\rbrace.
\end{equation}
In this experiment, we define the \textit{optimal constant cost} as the one with which the original highest quality is achieved in the shortest time, i.e. $\hat{c}~=~\argmin_{c\in C} t^{best}_c$.

\begin{table}[H]
\centering
{\small
\begin{tabular}[H]{|c|c|c|c|c|}
\hline 
\textbf{Problem} & $\hat{c}$ & $t^{best}_1$ & $t^{best}_{\hat{c}}$& $t^{best}_{\hat{c}}/t^{best}_1$\\ 
\hline 
SR &0.44 & 950000 & 225000 & 0.2368 \\ 
\hline 
WindFLO & 0.11 & 39 & 19 & 0.4871  \\ 
\hline 
Swimmer & 0.22 & 980000 & 516000 & 0.5265   \\ 
\hline 
Turbines & 0.33 & 3592 & 2112&  0.5880  \\ 
\hline 
\end{tabular} }
\caption{Optimal constant cost ($\hat{c}$) for each problem together with the time required to reach the original maximum quality ($t^{best}_{\hat{c}}$) and the percentage of the original time ($t^{best}_1$) that it supposes ($t^{best}_{\hat{c}}/t^{best}_1$).}	\label{TABLE_ConstantAnalysis}
\end{table}

The quality curves show the same pattern of behaviour for all problems. As the evaluation cost decreases, the quality curves improve the original curve, until reaching a cost at which the associated approximate function causes a worsening of quality in the solution. This cost is the optimal constant cost $\hat{c}$, with which the maximum quality is reached using less than $58.8\%$ of the time needed with the original cost in the worst case. In Symbolic Regressor, the approximate function $f_{0.44}$ allows us to reach the highest quality solution using only $23.68\%$ of the original time. Table~\ref{TABLE_ConstantAnalysis} shows in detail the optimal constant cost values and the times involved in reaching maximum quality. We observe that $\hat{c}$ is different in all problems. This shows that without the definition of a procedure such as the one proposed in Section \ref{SECTION_Method} we could not know what is the optimal evaluation cost in a new context, unless we first run the optimization algorithm with different cost values. Moreover, in the graphs associated with the Swimmer and Turbines, it can be seen that at an early stage of the execution process a lower than optimal evaluation costs perform better. The latter motivates the use of a changing evaluation cost during the optimization process.

\section{Applying \mbox{optecot} to benchmark and real-world problems}\label{SECTION_HeuristicApplication}
\begin{figure*}[!t]
\centering
\includegraphics[width=\textwidth]{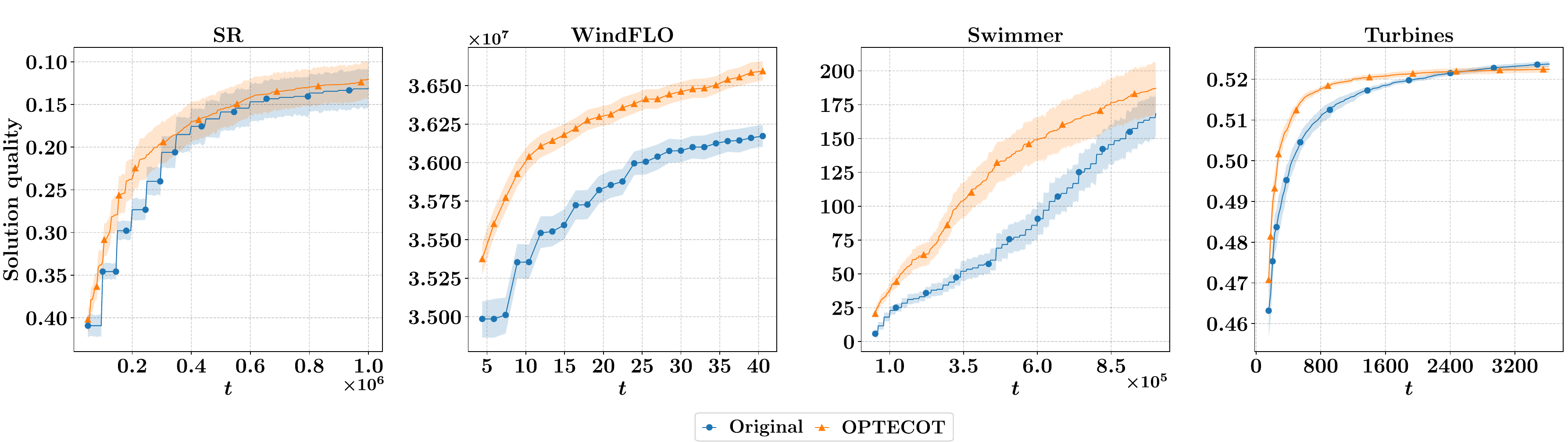}\setcounter{figure}{4}\caption{The quality evolution of the best solution found at each runtime by applying \mbox{OPTECOT} and using the original objective function to solve problems. The quality is measured in terms of the original objective function and the units of time in each problem are given in Table~\ref{TABLE_TrainTerms}.}\label{FIGURE_HeuristicApplicationQuality}
\end{figure*}
\begin{figure*}[!t]
\centering
\includegraphics[width=\textwidth]{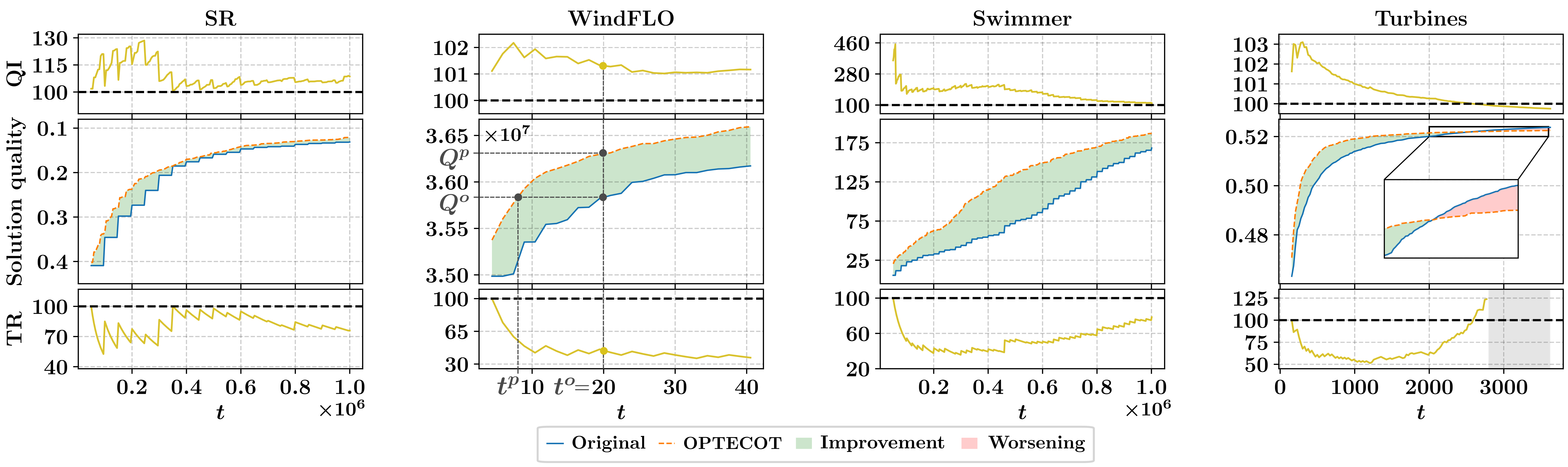}\setcounter{figure}{5}\caption{The mean quality evolution of the best solution found at each runtime by applying \mbox{OPTECOT} and using the original objective function (central graphs), together with the quality improvement curve (above) and the curve of the original time required (below). The mean quality is measured in terms of the original function, the quality improvement and time requirement are percentages, and the units of time in each problem are given in Table~\ref{TABLE_TrainTerms}.\protect\footnotemark}\label{FIGURE_HeuristicQualityTime}
\end{figure*} 
\footnotetext{In Turbines, the last part of the \mbox{OPTECOT} quality curve is below the original one. Consequently, as the original quality $Q^o$ is not reached with \mbox{OPTECOT} in the available runtime, the proposal reach times $t^p$ are unknown. For this reason, the yellow curve in the shaded grey area is not drawn.} 

After defining and testing the approximate objective functions in the previous section, in this section we apply \mbox{OPTECOT} to the described four problems. The experiments have been carried out using the hyperparameters shown in Table~\ref{TABLE_HeuristicParams}. The values of the parameters $|S|$ and $t^{period}$ are calculated using the Equations~\eqref{EQUATION_SampleSize} and \eqref{EQUATION_TimeFreq}. For the rest of the parameters instead, the values $\alpha=0.95$, $\beta=5$ and $\kappa=3$ have been empirically selected and correspond to those that work well in general for the four problems considered.

\begin{table}[H]
\centering
{\small
\begin{tabular}[H]{|c|c|c|c|c|c|}
\hline 
\textbf{Problem} & $\alpha$ & $\beta$ & $\kappa$& $|S|$ &$t^{period}$  \\ 
\hline 
SR & 0.95& 5 & 3 & 77 & 50000 \\ 
\hline 
WindFLO & 0.95& 5& 3&10 & 12.3   \\ 
\hline 
Swimmer  & 0.95& 5&3& 10 & 140000   \\ 
\hline 
Turbines  &0.95& 5&3& 10 & 516.9  \\ 
\hline 
\end{tabular} }
\caption{Parameter values for the \mbox{OPTECOT} definition.}\label{TABLE_HeuristicParams}
\end{table}

\subsection{Improvement of the solution quality} 
In Figure~\ref{FIGURE_HeuristicApplicationQuality} the original quality curves are shown together with the curves obtained by applying OPTECOT. We observe that \mbox{OPTECOT} improves the original solution quality. This improvement occurs in each problem during most of the execution process. Therefore, this corroborates that the time saving induced by the use of approximate functions compensates the minimal loss of accuracy (controlled by $\alpha$) and the time expend on the bisection method application.

The graphs in Figure~\ref{FIGURE_HeuristicQualityTime} illustrate in more detail the increase in quality and the time saved during the execution process by applying \mbox{OPTECOT}. For each problem, three graphs are drawn with the horizontal axis in common, corresponding to the algorithm execution time $t$. The central graph shows the area between the two mean quality curves drawn in Figure \ref{FIGURE_HeuristicApplicationQuality} for each problem. The green area indicates that the difference obtained after applying \mbox{OPTECOT} results in an improvement, while the red area represents a worsening. These areas are interpreted in terms of the \textit{quality increase} (QI) and the original \textit{time required} (TR). The height of the curve above the central graph corresponds to the percentage of quality improvement obtained by using \mbox{OPTECOT} for each execution time (its associated height at a time of 20 seconds in WindFLO is exactly $(Q^p/Q^o)\cdot 100$). The yellow curve at the bottom instead, indicates the percentage of time required by \mbox{OPTECOT} to reach the original quality at each time (at a time of 20 seconds in WindFLO the height of this curve is $(t^p/t^o)\cdot 100$), e.g. a 50\% means that \mbox{OPTECOT} needs half of the time to reach the same solution quality as using the original objective function.

Generally, we observe that the quality and time curves exceed and are below 100 during the whole execution process respectively, except in the last runtimes of Turbines. In addition, the greatest improvements occur from the beginning until a moderate portion of the available runtime has elapsed, and it tends to decrease with $t$. This shows the capability of \mbox{OPTECOT} to improve the solution quality in the optimization process when the high evaluation cost makes the computational budget insufficient for the algorithm to converge. In this case, \mbox{OPTECOT} provides its maximum performance benefit, as shown by the reduction in the required time to reach the original quality. The results obtained for WindFLO and Swimmer are noteworthy (the associated runtime limits seem to be far from the convergence), where an average of $45.65\%$ and $53.25\%$ of the original time is used during the whole execution process, respectively. In contrast, in the case of Turbines, the proposed value for the unavailable $t^{max}$ runtime limit proved to be enough for the optimization algorithm to converge. Consequently, \mbox{OPTECOT} only provided a benefit in the first $69.93\%$ of the computation time. This example shows how in cases where the available budget is large enough, it is better to use the original objective function directly.
\vspace{-1mm}
\subsection{Real-time cost adjustment} 
\begin{figure}[H]
\centering
\includegraphics[width=8.3cm]{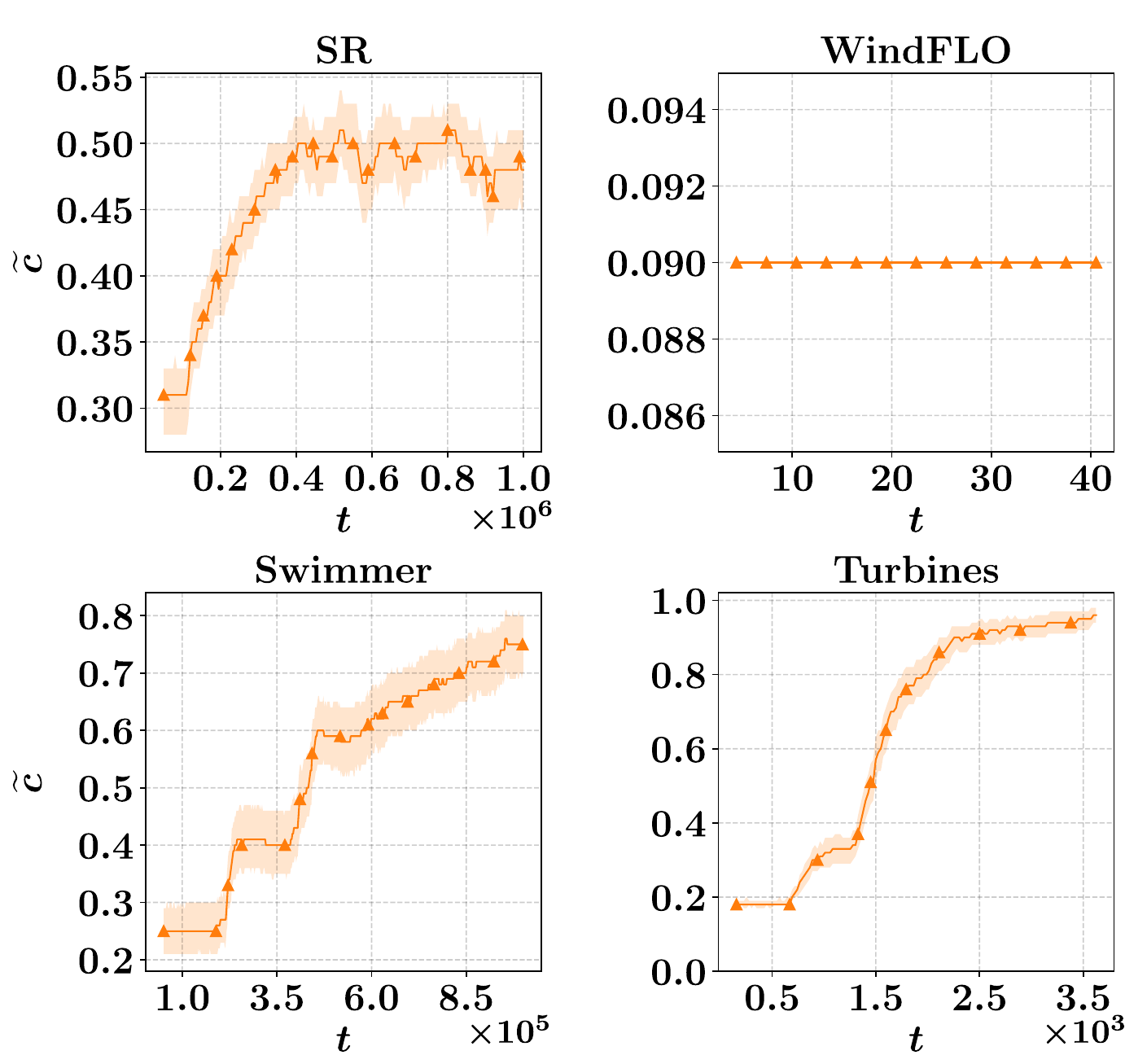}\caption{The evolution of the optimal cost defined by \mbox{OPTECOT} at each runtime. The units of time in each problem are given in Table~\ref{TABLE_TrainTerms}.}\label{FIGURE_HeuristicApplicationCost}
\end{figure}

The curves in Figure~\ref{FIGURE_HeuristicApplicationCost} represent the mean optimal evaluation cost considered by OPTECOT in each time of the optimization algorithm execution. In Symbolic Regressor, Swimmer and Turbines they are neither constant nor monotonic, which corroborates the need to consider the use of approximate functions of different costs during the solution process. The downward slopes of these curves allow us to choose a lower cost function saving unnecessary costs, and the upward slopes show the need to use a higher cost approximation to guarantee the correct ranking of the solutions that form the population. In WindFLO instead, the optimal evaluation cost is constant and significantly lower than in the rest problems (as shown in Table~\ref{TABLE_ConstantAnalysis}). This is because in the readjustments the bisection method has always provided the same approximate value for the optimal cost. The optimal costs are too small to be reached in the iterations available for bisection, becoming the bisection invariant in every iteration. Finally, we observe that overall the cost curves have an ascending trend with the runtime. This seems to indicate that as the optimization process advances, approximations of higher cost and accuracy are required. In any case, the results obtained show that the proposed method can adjust the cost in real-time whatever its optimal behaviour is during the execution process.
\vspace{-4mm}

\section{Limitations}\label{SECTION_Limitations}
This section summarizes the limitations of the proposed method. \mbox{OPTECOT} is designed to solve a computationally expensive black-box optimization problem using RBEAs with insufficient budget. For this purpose, evaluation time is reduced by estimating the cost-indexed parameter value that provides an approximation with reasonable balance between the evaluation cost and its accuracy. Therefore, the application of the method has the following limitations:
\begin{itemize}
\item[$(A)$] \textit{The use of RBEAs as optimization algorithm.} The algorithm used to solve the optimization problem must work with a set of solutions that will be updated iteratively considering their ranking in the selection step (for more details see Section~\ref{SUBSECTION_OptimizationAlgorithm} where RBEAs are defined).
\item[$(B)$]\textit{High evaluation cost.} The evaluation cost of the objective function must be a bottleneck. It has to be large enough so that the number of evaluations allowed by the budget is insufficient for the population-based evolutionary algorithm to provide a quality solution.
\item[$(C)$]\textit{Cost-indexed parameter-based approximations.} The objective function must allow the definition of an approximation with a parameter that regulates its computational cost. Moreover, the approximations with lower evaluation costs should have lower accuracy, and vice versa.
\end{itemize}

\section{Discussion and Future Work}\label{SECTION_DisscussionLimitationsFuturework}
\begin{figure*}[!t]

	\begin{subfigure}[b]{0.3\textwidth}
		\includegraphics[height=5cm]{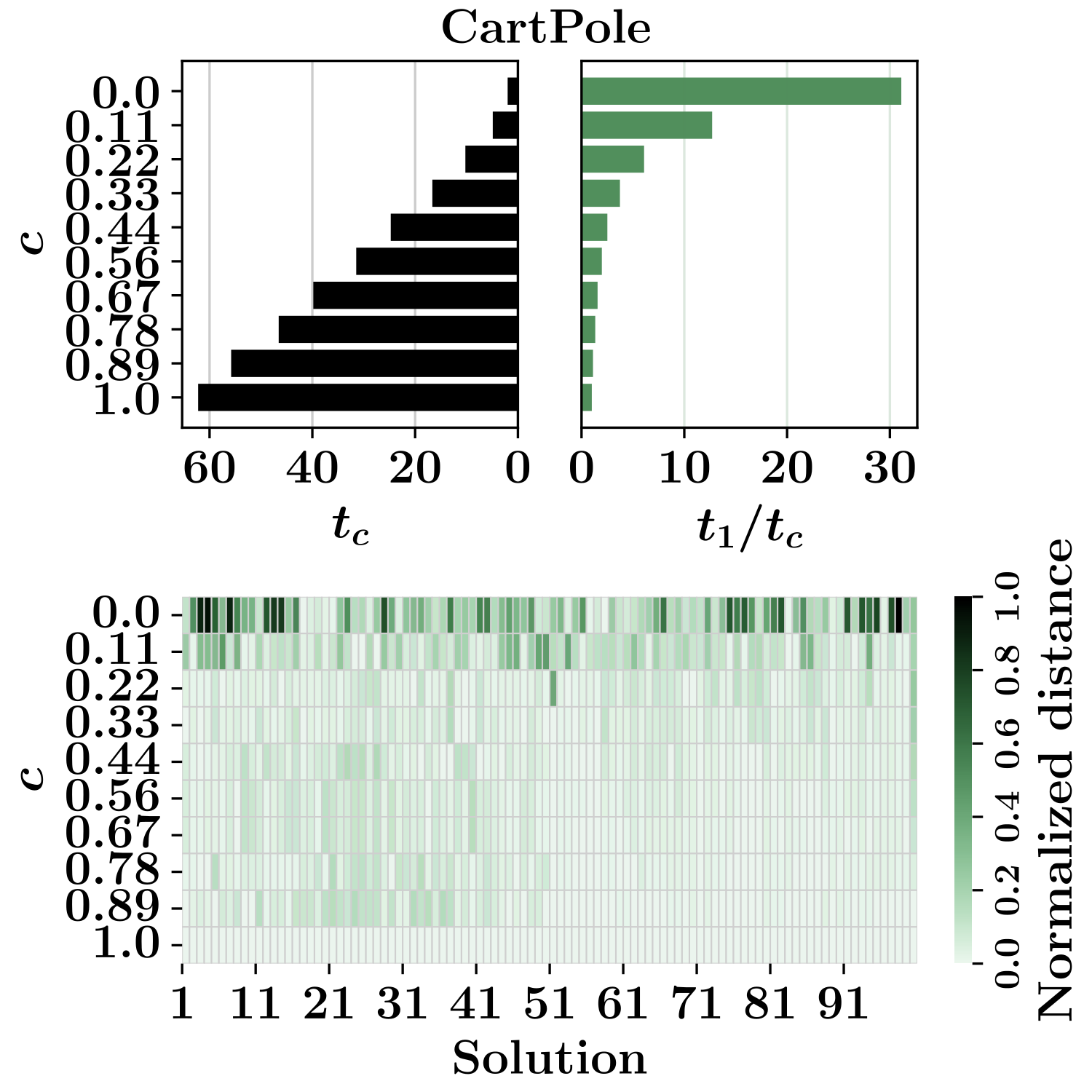}
    	\caption{Evaluation times, extra evaluations and ranking preservation for different costs.}\label{SUBFIGURE_sutability}
	\end{subfigure}
	\hspace{2mm}
	\begin{subfigure}[b]{0.28\textwidth}
    	\includegraphics[height=5cm]{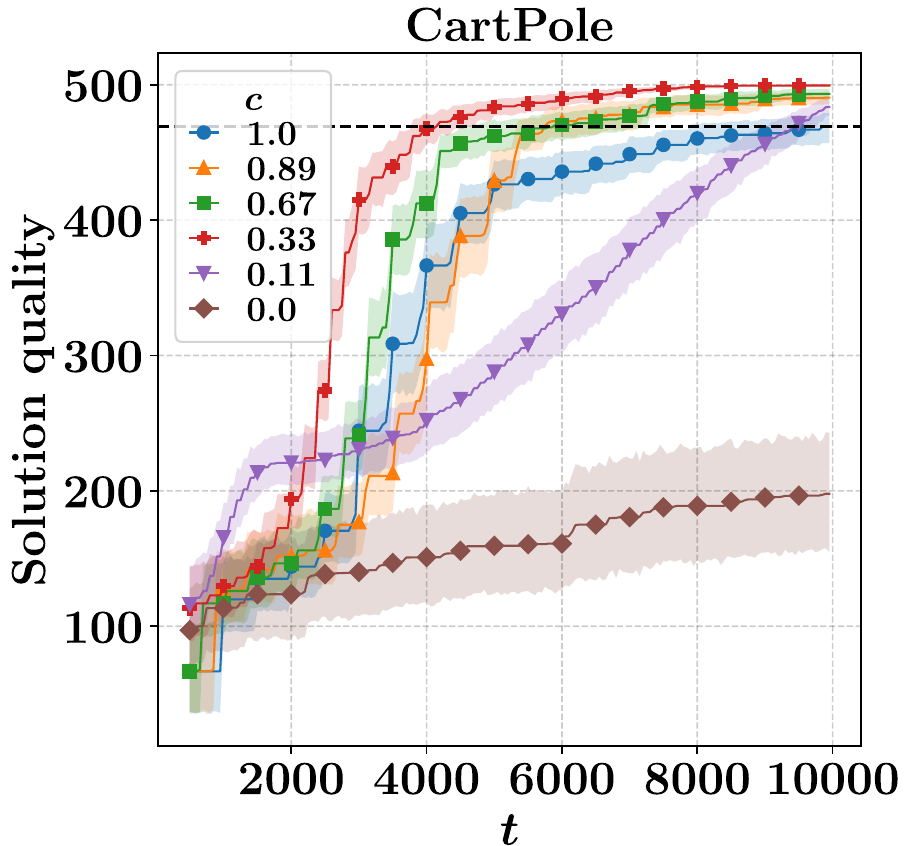}
    	\caption{Quality curves obtained by running PPO using different constant costs.}\label{SUBFIGURE_effectiveness}
    	\label{fig:second}
	\end{subfigure}
	\hspace{5mm}
	\begin{subfigure}[b]{0.35\textwidth}	

    \begin{subfigure}[t]{0.45\textwidth}
    	\centering
    	
    	\begin{tabular}{|c|c|}
    		\hline 
    		$x$ & Policy \\ 
    		\hline 
    		$f$ & Reward \\ 
    		\hline 
    		$\theta$ & Time-step \\ 
    		\hline 
    		$\theta_0$ & 0.2 \\ 
    		\hline 
    		$\theta_1$ & 0.02 \\ 
    		\hline 
		\end{tabular}
	\end{subfigure}
	\hspace{-2mm}
	 \begin{subfigure}[t]{0.45\textwidth}
    	\centering
		
		\begin{tabular}{|c|c|}
			\hline
     		Algorithm  & PPO\\ 
    		\hline 
    	    $t$ & Steps \\ 
			\hline 
    		$t^{max}$ & 10000 \\ 
    		\hline 
    		$\hat{c}$ & 0.33 \\ 
    		\hline 
    		$t_1^{best}$ & 9850 \\ 
    		\hline 
    		$t_{\hat{c}}^{best}$  & 4150 \\ 
    		\hline 
    		$t_{\hat{c}}^{best}/t_1^{best}$ & 0.4213 \\ 
    		\hline 
    	\end{tabular}
		\vspace{7mm}
	\end{subfigure}
    	\caption{On the left, optimization problem and cost-indexed parameter. On the right, terms related to train process and optimal constant cost \cite{farama-foundation_gymnasium_2023,raffin_stable_2023-1}.}\label{SUBFIGURE_tables}
	\end{subfigure}
       
\caption{Experimental results obtained with CartPole problem using PPO algorithm.}\label{FIGURE_CartPole}
\end{figure*}

In this section, we discuss the limitations summarized in Section~\ref{SECTION_Limitations}, in addition to an additional issue. Following the order in which the previous limitations have been presented, Sections~\ref{SUBSECTION_discussionA} and \ref{SUBSECTION_discussionB} discuss the opposite cases to limitations $(A)$ and $(B)$ respectively, and Section \ref{SUBSECTION_discussionC} argues possible strategies to improve the efficiency of the bisection method that plays an essential role in the selection of the cost-indexed parameter value with which the approximations are defined (limitation $(C)$). Sections~\ref{SUBSECTION_discussionD} instead, addresses the additional topics of manual cost tuning. These discussions are supported by the results obtained in Section~\ref{SECTION_HeuristicApplication} where we showed the advantages of using \mbox{OPTECOT} with different algorithms and problems in terms of solution quality and time.

\subsection{OPTECOT beyond RBEAs}\label{SUBSECTION_discussionA}

RBEAs are population-based optimization algorithms based on the relative performance of solutions within a population, expressed in terms of rankings. Note that the experiments have been carried out using two kinds of RBEAs: GP \cite{ahvanooey_survey_2019} and CMA-ES \cite{auger_tutorial_2012}. However, \mbox{OPTECOT} could be adapted to be applied to other types of algorithms. 

Proximal Policy Optimization (PPO) \cite{raffin_ppo_nodate-1} is a popular policy gradient method for reinforcement learning, which alternates between sampling data through interaction with the environment, and optimizing an approximate objective function using stochastic gradient ascent \cite{schulman_proximal_2017}. Since it is not an RBEA, it is not directly compatible with \mbox{OPTECOT}, although it is possible to analyze the suitability and effectiveness of a set of approximate functions as has been done in Section~\ref{SECTION_PreliminaryExperiments}. As an example, the results of the initial analysis mentioned above for the CartPole Classic Control problem \cite{barto_neuronlike_1983} using the PPO algorithm are shown in Figure~\ref{FIGURE_CartPole}, in which we selected the time-step parameter to define approximate functions (see the first table in Figure~\ref{SUBFIGURE_tables} for more detail). 

Figures \ref{SUBFIGURE_sutability} shows that the evaluation time is strictly increasing with respect to cost. In addition, the ranking of the solution sample is preserved until relatively lower cost values are reached. Therefore, the selected parameter is a good candidate to define the approximate functions of different costs. On the other hand, Figure \ref{SUBFIGURE_effectiveness} shows that with the approximate function $f_{0.33}$ the same maximum quality is achieved as with the original objective function, using only $42.13\%$ of the time. Moreover, although considering a constant cost $f_{0.33}$ is the most beneficial approximation, observe that at the beginning of the execution process lower cost approximations obtain higher quality results, such as $f_{0.11}$. In conclusion, the obtained results show that it is very likely that the use of lower-cost approximate functions with variable cost values during the execution of the algorithm leads to an improvement in the solution quality in a given runtime for this problem. This suggests that a generalization of the proposed method for other optimization algorithms that are not RBEAs might be possible.

\subsection{False convergence of approximate functions}\label{SUBSECTION_discussionB}
When the evaluation cost is not a bottleneck and a large enough computational budget is available to solve an optimization problem, the original objective function should be used. Applying \mbox{OPTECOT} in such cases does not make sense, since there is no need to save time in the evaluation process and the use of approximate functions may result in premature convergence to suboptimal solutions. This phenomenon can be related to the false convergence in surrogate models \cite{tong_surrogate_2021,sykulski_considerations_2007,zhou_combining_2007,jones_taxonomy_2001}, when the provided solution is the optima of the surrogate, but not of the original objective function. A similar situation has been observed for the final runtimes in Turbines, where the selected $t^{max}$ allows to solve the problem with the original function. However, \mbox{OPTECOT} is specifically designed to address the optimization challenge effectively in situations where computational resources are insufficient to achieve convergence. In such case, the problem of false convergence does not apply.

\subsection{Bisection effectiveness adjusting low optimal costs} \label{SUBSECTION_discussionC}

The bisection procedure to approximate the optimal cost described in Section~\ref{SUBSECTION_bisection} iteratively divides the search interval in half. This way of dividing implies that in cases where the optimal cost is close to the maximum and minimum values, the effectiveness of the bisection can be improved. In that case, the iterative reduction of the interval size is too small, and the number of iterations considered are not enough to reach the optimal cost. This limitation does not affect to the case in which the optimal cost is close to the maximum, thanks to the interruption criterion applied in \mbox{OPTECOT} (see Section~\ref{SUBSECTION_OPTECOT} for details). However, for optimal costs close to the minimum, the limitation is still present (an example of this is the already mentioned case of WindFLO). As future work, it would be possible to modify the bisection algorithm described in Section~\ref{SUBSECTION_bisection}. The way of dividing the cost interval could be defined after assessing what is the increase in the approximation accuracy concerning the previous iteration. This accuracy increase could be used as a guide to determine where to split the interval (in the middle or closer to one of the extremes) and define the new cost to be considered as the splitting point.

\subsection{\mbox{OPTECOT} versus manual cost tuning}\label{SUBSECTION_discussionD}

In the experimental Section~\ref{SECTION_PreliminaryExperiments}, before applying \mbox{OPTECOT} to the selected problems, we analyze the performance when using constant cost approximations. By comparing the results of the two experiments, we observe that sometimes manual cost tuning can provide better results than using \mbox{OPTECOT}. For example, in Figures~\ref{FIGURE_ConstantEvaluationCost} and \ref{FIGURE_HeuristicApplicationQuality} we observe that the constant costs $0.44$ and $0.22$ in the Symbolic Regressor and Swimmer environments respectively achieve better results than applying \mbox{OPTECOT}. However, the optimal constant cost value depends on the problem characteristics, and it can only be calculated by running the algorithm with several costs, which is a much more expensive procedure than solving the optimization problem directly without cost reduction procedures. 

Alternatively, a small portion of the available execution time could be reserved to run the algorithm using approximations of different costs and select the optimal constant cost. Then, the remaining time could be used to solve the problem with the selected cost. The drawback of this alternative is that at an initial stage, the optimal constant cost may be lower than the real optimal constant cost. For instance, in Figure~\ref{FIGURE_ConstantEvaluationCost} for Swimmer and Turbines, it can be seen that at an early stage of the execution process, lower than optimal constant evaluation costs perform better. Consequently, the solution process would result in an inaccurate exploration, since although more evaluations could be done (optimal constant cost too small), the accuracy would not be appropriate (even if a small cost works well at the beginning, it will lose accuracy as the algorithm converges). 

In conclusion, unlike manual cost tuning, \mbox{OPTECOT} allows evaluating additional solutions in the same amount of time by controlling the evaluation accuracy in a context where the available optimization time budget is low.

\section{Conclusion}\label{SECTION_Conclusions}
In this work, we have presented \mbox{OPTECOT}, a technique to speed up the optimization process of population-based evolutionary algorithms that rely on the relative performance of solutions within a population, expressed in rankings. This method is particularly suitable for computationally expensive black-box optimization problems when the available computational budget is insufficient for the algorithm to converge. \mbox{OPTECOT} operates on a set of approximate objective functions with different accuracies obtained by modifying a parameter that regulates their computational cost. The procedure consists in evaluating the populations using the approximation with the optimal cost, which finds the trade-off between saving evaluation time and maintaining enough accurate population ranking. A notable feature of \mbox{OPTECOT} is its ability to automatically determine when to update the optimal cost of the approximate function and efficiently calculate this cost.

The effectiveness of the proposal has been demonstrated using two population-based evolutionary algorithms in four problems: a regression problem, two fluid simulation problems and a policy search problem. The improvements are particularly relevant when the available computational budget does not allow to solve the problem using the original objective function. In fact, on average, it has taken less than half of the original time to reach the original solution qualities.

Alongside the paper, we provide the source code available in our GitHub repository: \url{https://github.com/JudithEtxebarrieta/OPTECOT}. This code allows the reproduction of the experiments, in addition to being useful as a guide to apply \mbox{OPTECOT} on new problems.

\section*{Acknowledgments}
This research is supported by Basque Government through the Elkartek program under the project KONFLOT KK-2022/00100 and the BERC 2022-2025 program, and
by the Ministry of Science and Innovation: BCAM Severo Ochoa accreditation CEX2021-001142-S/ MICIN/ AEI/ 10.13039/ 501100011033.

\appendix
\subsection{Confidence intervals of the populations variances}\label{APPENDIX_CI}
The designed procedure defines when to readjust the cost of the approximate objective function after assessing if the variance of the scores of the last population is significantly different from the previous variances. A larger variance facilitates ordering correctly solutions using a lower cost approximation (less accurate), while a smaller variance requires a higher cost approximation (more accurate) to guarantee that the solutions are correctly ranked. Therefore, monitoring the behaviour of the variance from population to population allows us to guess when the approximation accuracy might not be adequate.

The condition that we have designed for the definition of a ``significant variance change" is inspired by the works focused on \textit{concept drift} adaptation \cite{gama_survey_2014}. This phenomenon appears when the data streams available to fit a predictive model present distribution changes over time, arising the need to adjust the model as new data are collected to ensure that the training data correctly represent the data to be predicted. Some algorithms in the literature about concept drift monitor the predictive error of the current model during the prediction process \cite{hutchison_learning_2004,klinkenberg_adaptive_nodate,lu_learning_2018}. When it is above the deviating mean of the previous errors, the current model is considered to be obsolete and it is retrained with the most recent data. In our case instead, the accuracy of an approximation may be affected when a lower or higher variance is detected. Therefore,
we adapt the definition saying that the approximation must be readjusted when the last recorded variance is not in the following interval
\begin{equation}
\CI =\mean\big(\lbrace v_1,...,v_{\beta}\rbrace\big)\pm 2\cdot\std\big(\lbrace v_1,...,v_{\beta}\rbrace\big)
\end{equation}
where $\lbrace v_1,...,v_{\beta}\rbrace$ is the set formed by the previous $\beta$ variances, and $\mean(\cdot)$ and $\std(\cdot)$ represent the mean and standard deviation respectively. This interval is named as confidence interval in Section \ref{SUBSECTION_OPTECOT}. The factor of the standard deviation is the critical value of the interval, which in the case of $95\%$ confidence interval is $2$.

\subsection{Relationship between cost, evaluation time and parameter}\label{APPENDIX_RelationCostParamaeter}

This appendix defines the explicit relationship considered between the cost $c$, the evaluation time $t_c$ and the cost-indexed parameter $\theta_c$, which are necessary for the reproducibility of the experiments. Although \mbox{OPTECOT} has been defined in terms of $c$, in practice only the relation between $t_c$ and $\theta_c$ is used. To link the theoretical and practical concepts, we first relate $t_c$ to $c$ and then relate $c$ to $\theta_c$. Combining both, the explicit relation between $t_c$ and $\theta_c$ is obtained, which can be defined in any problem that has a parameter capable of controlling the evaluation time monotonically.

\subsubsection{From evaluation time to cost}
\begin{figure}[H]
\centering
\includegraphics[width=\columnwidth]{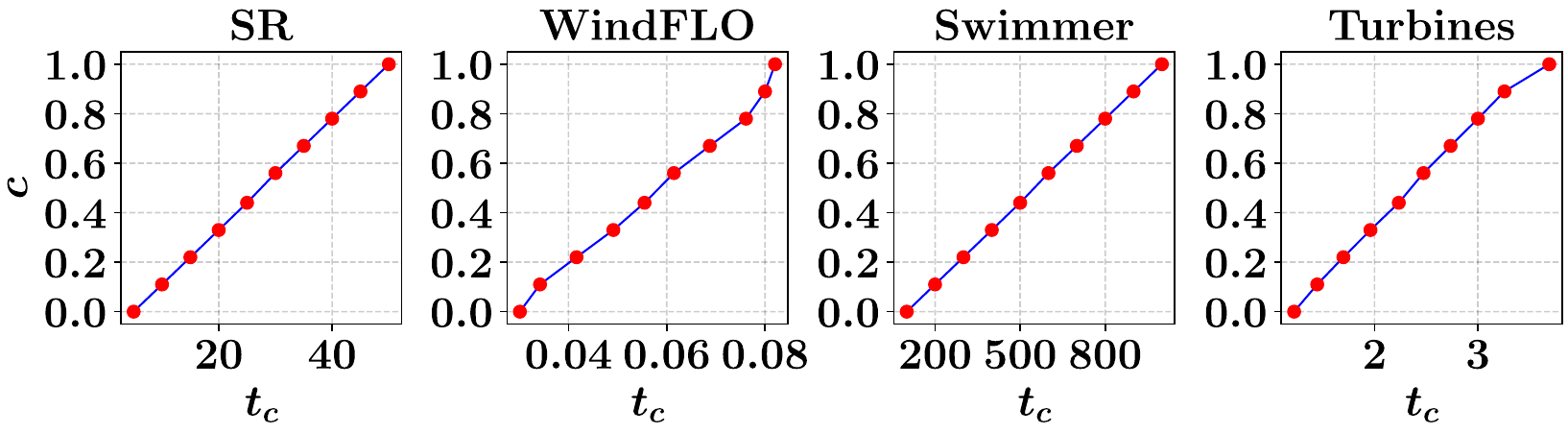}\caption{Relationship between cost and evaluation time.}\label{FIGURE_Interpolations}
\end{figure}

For the practical application of \mbox{OPTECOT}, the bisection is applied to the evaluation times instead of to the costs (which are its abstraction). Therefore, for reproducibility, the relationship between a cost $c$ and its associated evaluation time $t_c$ is necessary. Its ideal definition has been given in Section \ref{SECTION_FormalDefinition} by Equation~\eqref{EQUATION_Cost}, but as we have already mentioned, in practice we use an approximation. We approximate Equation~\eqref{EQUATION_Cost} using the average evaluation times calculated in Section \ref{SUBSECTION_ResultsSuitability} after evaluating the random set of solutions, and by interpolation we estimate a linear expression defining the relation. Let be $\big\lbrace(t_c,c)\big\rbrace_{c\in C}$ the set formed by the pairs of costs and the associated average times per evaluation calculated in Section~\ref{SUBSECTION_ResultsSuitability}. The relationship between $t_c$ and $c$ for any $t_c\in [t_0,t_1]$ is defined as
\begin{equation}\label{EQUATION_Interpolation}
c=c'+\dfrac{c''-c'}{t_{c''}-t_{c'}}(t_c-t_{c'})
\end{equation}
where $t_{c'}$ and $t_{c''}$ are the closest evaluation times below and above $t_c$ among the 10 possible times $\lbrace t_c\rbrace_{c\in C}$. Figure~\ref{FIGURE_Interpolations} shows graphically the linear relationship \eqref{EQUATION_Interpolation} obtained for each problem, where the red dots are associated with the known coordinates and the blue polygonal curve represents the interpolation.

\subsubsection{From cost to cost-indexed parameter}
In this work, the approximate functions have been presented as lower-cost approximations of the original objective function obtained by modifying an original value of a strategic parameter. As the objective of this work is to reduce the cost of the RBEA while maintaining good solution quality, this is the main reason why in our proposal the approximations have been defined only in terms of cost, although they are actually defined with an explicit value of the parameter. In the experiments, the definition of the relationship between the cost and the value of the $\theta$ parameter has been omitted for simplicity. For reproducibility instead, it is necessary to know this relation to have the exact definition of the approximate function associated with a cost. This allows to make the pertinent evaluations during the execution of the algorithm.

In practice, we modify the accuracy of the original parameter value $\theta_1$ to define approximate functions. It is linearly related to the cost, less accurate parameter values give us an approximation of lower cost. Formally, for each cost $c\in [0,1]$ the value of the parameter considered is
\begin{equation}\label{EQUATION_RelationCostParam}
\theta_c=
\begin{cases}
\theta_1\cdot a_c, &\text{if}~\theta_1>\theta_0\\
\theta_1/ a_c, &\text{otherwise}
\end{cases}
\end{equation}
where $a_c = a_0 +c\cdot(a_1 −a_0)$ is the parameter accuracy linearly related to the cost. Note that $a_0$ takes the values $\theta_0/\theta_1$ and $\theta_1/\theta_0$ in the cases that $\theta_1 > \theta_0$ and $\theta_1 < \theta_0$ respectively, while $a_1 = 1$ in both cases. 

\bibliographystyle{IEEEtran}
\bibliography{main.bib} 

%\listoftables
%\listoffigures

\vfill

\end{document}